\documentclass[final]{cvpr}

\usepackage{times}
\usepackage{epsfig}
\usepackage{graphicx}
\usepackage{amsmath}
\usepackage{amssymb}
\usepackage{color,xcolor}
\usepackage{multirow}
\usepackage{enumerate}
\usepackage{algorithm}
\usepackage{algorithmic}
\usepackage{rotating}
\usepackage{booktabs} 
\usepackage{xparse}
\usepackage[title]{appendix}

\newtheorem{theorem}{Theorem}
\newtheorem{proof}{Proof}

\newcommand{\heading}[1]{\noindent\textbf{#1}}

\newcommand{\textRed}[1]{\textcolor[rgb]{0.75,0,0}{#1}}
\newcommand{\textGreen}[1]{\textcolor[rgb]{0.33,0.51,0.21}{#1}}

\def\ie{\textit{i.e.}}
\def\eg{\textit{e.g.}}
\def\etal{\textit{et~al.}}
\definecolor{mypink}{RGB}{219, 48, 122}

\newcommand\appref[1]{Appendix.~\ref{#1}} 

\usepackage[pagebackref=true,breaklinks=true,colorlinks,bookmarks=false]{hyperref}

\colorlet{dark-blue}{blue!70!black}
\hypersetup{
    colorlinks=true,%
    citecolor=dark-blue,%
    filecolor=dark-blue,%
    linkcolor=red,%
    urlcolor=magenta
}

\def\cvprPaperID{815} 
\def\confYear{CVPR 2021}

\begin{document}


\title{Joint Noise-Tolerant Learning and Meta Camera Shift Adaptation \\ for Unsupervised Person Re-Identification}

\author{Fengxiang Yang$^{\textcolor{mypink}{1}}\thanks{Equal contribution: yangfx@stu.xmu.edu.cn}$~, Zhun Zhong$^{\textcolor{mypink}{2}*}$, Zhiming Luo$^{\textcolor{mypink}{3}}\thanks{Corresponding author: \{zhiming.luo, szlig\}@xmu.edu.cn}$~, Yuanzheng Cai$^{\textcolor{mypink}{4}}$, Yaojin Lin$^{\textcolor{mypink}{5}}$, Shaozi Li$^{\textcolor{mypink}{1}\dag}$, Nicu Sebe$^{\textcolor{mypink}{2}}$ \\
 \small{\textcolor{mypink}{1} Artificial Intelligence Department, Xiamen University} \\
 \small{\textcolor{mypink}{2} Department of Information Engineering and Computer Science, University of Trento}\\ 
 \small{\textcolor{mypink}{3} Postdoc Center of Information and Communication Engineering, Xiamen University} \\
 \small{\small{\textcolor{mypink}{4} Minjiang University} \space \space \textcolor{mypink}{5} Minnan Normal University} \\
 \small{Project: \url{https://github.com/FlyingRoastDuck/MetaCam_DSCE}}
}

\maketitle

\begin{abstract}

This paper considers the problem of unsupervised person re-identification (re-ID), which aims to learn discriminative models with unlabeled data. One popular method is to obtain pseudo-label by clustering and use them to optimize the model. Although this kind of approach has shown promising accuracy, it is hampered by 1) noisy labels produced by clustering and 2) feature variations caused by camera shift. The former will lead to incorrect optimization and thus hinders the model accuracy. The latter will result in assigning the intra-class samples of different cameras to different pseudo-label, making the model sensitive to camera variations. In this paper, we propose a unified framework to solve both problems. Concretely, we propose a Dynamic and Symmetric Cross-Entropy loss (DSCE) to deal with noisy samples and a camera-aware meta-learning algorithm (MetaCam) to adapt camera shift. DSCE can alleviate the negative effects of noisy samples and accommodate to the change of clusters after each clustering step. MetaCam simulates cross-camera constraint by splitting the training data into meta-train and meta-test based on camera IDs. With the interacted gradient from meta-train and meta-test, the model is enforced to learn camera-invariant features. Extensive experiments on three re-ID benchmarks show the effectiveness and the complementary of the proposed DSCE and MetaCam. Our method outperforms the state-of-the-art methods on both fully unsupervised re-ID and unsupervised domain adaptive re-ID. 

\end{abstract}

\section{Introduction}
\begin{figure}[!t]
    \centering
    \includegraphics[width=0.98\linewidth]{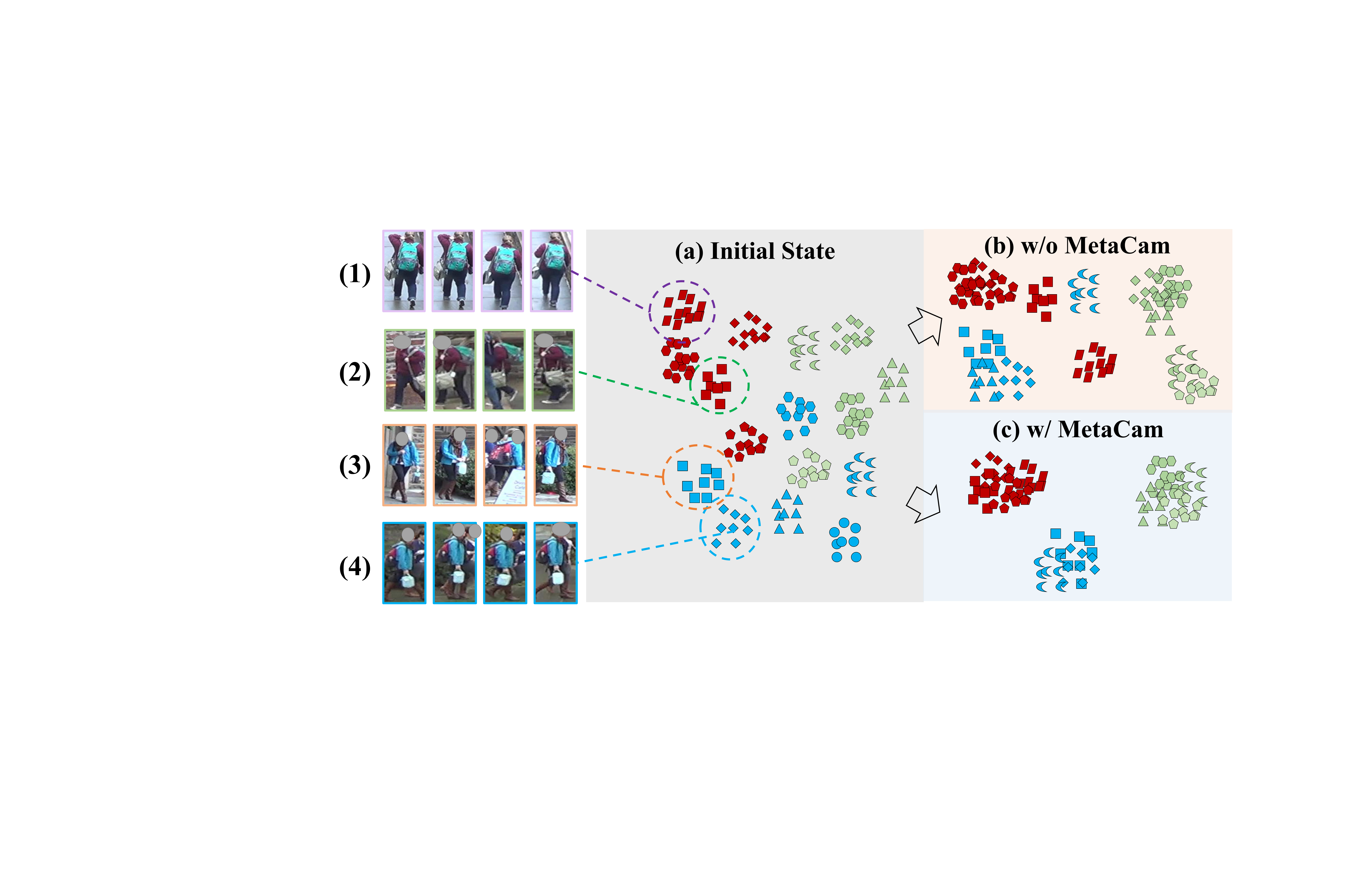}
    \vspace{-.06in}
    \caption{Illustration of camera variations in person re-ID (a) and the comparison between methods trained without or with the proposed MetaCam ((b) and (c), respectively). Different colors represent different identities and different shapes indicate different camera IDs. At the initial state, samples under different cameras may suffer from appearance changes of viewpoints ((1) \& (2)), illumination ((3) \& (4)), and other factors. Without considering this factor, the trained model may be sensitive to camera variations and may wrongly split intra-class features to different centers. Our proposed MetaCam enables the model to learn camera-invariant features by explicitly considering cross-camera constraint.}
    \label{fig:metacam} 
    \vspace{-.1in}
\end{figure}

Person re-identification (re-ID) attempts to find matched pedestrians of a query in a non-overlapping camera system. Recent CNN-based works~\cite{sun2018beyond, wang2018learning} have achieved impressive accuracies, but their success is largely dependent on sufficient annotated data that require a lot of labeling cost. In contrast, it is relatively easy to obtain a large collection of unlabeled person images, fostering the study of unsupervised re-ID. Commonly, unsupervised re-ID can be divided into two categories depending on whether using an extra labeled data, \textit{i.e.}, unsupervised domain adaptation (UDA)~\cite{PTGAN, zhong2019invariance, ge2020mutual} and fully unsupervised re-ID (FU)~\cite{lin2019bottom, lin2020unsupervised, zeng2020hierarchical}. In UDA, we are given a labeled source domain and an unlabeled target domain. The data of two domains have different distributions and are used to train a model that generalizes well on the target domain. The fully unsupervised re-ID is more challenging since only unlabeled images are provided to train a deep model. In this study, we will mainly focus on this setting, and call it as unsupervised re-ID for simplicity.

Recent popular unsupervised re-ID methods~\cite{lin2019bottom, wang2020unsupervised, lin2020unsupervised, zeng2020hierarchical} mainly adopt clustering to generate pseudo-label for unlabeled samples, enabling the training of the model in a supervised manner. Pseudo-label generation and model training are applied iteratively to train an accurate deep model. Despite their effectiveness, existing methods often ignore two important factors during this process. (1) \textit{Noisy labels brought by clustering}. The clustering algorithm cannot ensure intra-samples to be assigned with the same identity, which inevitably will introduce noisy labels in the labeling step. The errors of noisy labels will be accumulated during training, thereby hindering the model accuracy. (2) \textit{Feature variations caused by camera shifts}. As shown in Fig.~\ref{fig:metacam}, intra-class samples under different cameras may suffer from  the changes of viewpoint (\eg, (1) and (2) in Fig.~\ref{fig:metacam}), illumination (\eg, (3) and (4) in Fig.~\ref{fig:metacam}), and other environmental factors. At the start of unsupervised learning (``initial state'' in Fig.~\ref{fig:metacam}), these significant variations will cause large gaps between the intra-class features of different cameras. In such a situation, it is difficult for the clustering algorithm to cluster samples with the same identity from all cameras into the same cluster. Consequently, training with the samples mined by the clustering will lead to unexpected separation for intra-class samples (``w/o MetaCam'' in Fig.~\ref{fig:metacam}) and the model might be sensitive to camera variations during testing. In this paper, we attempt to solve the above two crucial problems for robust unsupervised re-ID.

\textit{For the first issue}, we try to adopt the technique of learning with noisy labels (LNL) for robust training. LNL is well-studied in image classification, however, most of the existing methods cannot be directly applied to our scenario. This is because the centers and pseudo-label will change after each clustering step. To overcome this difficulty, this paper proposes a \textit{dynamic and symmetric cross-entropy loss} (DSCE) for unsupervised re-ID. We maintain a feature memory to store all image features, which enables us to dynamically build new class centers and thus to be adaptable to the change of clusters. With the dynamic centers, a robust loss function is proposed for mitigating the negative effects caused by noisy samples.

\textit{For the second issue}, we attempt to explicitly consider camera-invariant constraint during training. Indeed, person re-ID is a cross-camera retrieval process, aiming to learn a model that can well discriminate samples under different cameras. If a model trained with samples from some of the cameras can also generalize to distinguish samples from the rest of the cameras, then, we could obtain a model that can extract the intrinsic feature without camera-specific bias and is robust to camera changes. Inspired by this, this paper introduces a camera-aware meta-learning (MetaCam), which aims to learn camera-invariant representations by simulating the cross-camera re-identification process during training. Specifically, MetaCam separates the training data into meta-train and meta-test, ensuring that they belong to entirely different cameras. We then enforce the model to learn camera-invariant features under both camera settings by updating the model with meta-train and validating the updated model with meta-test. Along with learning from different meta divisions, the model is gradually optimized to generalize well under all cameras. As shown in Fig.~\ref{fig:metacam}, MetaCam gathers intra-class features of different cameras into the same cluster, which is beneficial for mining pseudo-label and learning camera-invariant features. In summary, our main contributions can be summarized in three aspects:

\begin{itemize}
    \vspace{-.07in}
    \item We propose a dynamic and symmetric loss (DSCE), which enables the model to be robust to noisy labels during training in the context of changes of clusters and thus promotes the model performance.
    \vspace{-.07in}
    
    \item We propose a camera-aware meta-learning algorithm (MetaCam) for adapting the shifts caused by cameras. By simulating the cross-camera searching process during training, MetaCam can effectively improve the robustness of the model to camera variations.
    \vspace{-.07in}
    \item We introduce a unified framework that can jointly take advantage of the proposed DSCE and MetaCam, enabling us to learn a more robust re-ID model.
    \vspace{-.07in}
\end{itemize}

Extensive experiments on three large-scale datasets demonstrate the advantages of our DSCE and MetaCam for the fully unsupervised re-ID. 
Besides, further experiments on the UDA setting show that our method can also achieve state of the art.

\begin{figure*}[!t]
    \centering
    \includegraphics[width=0.99\linewidth]{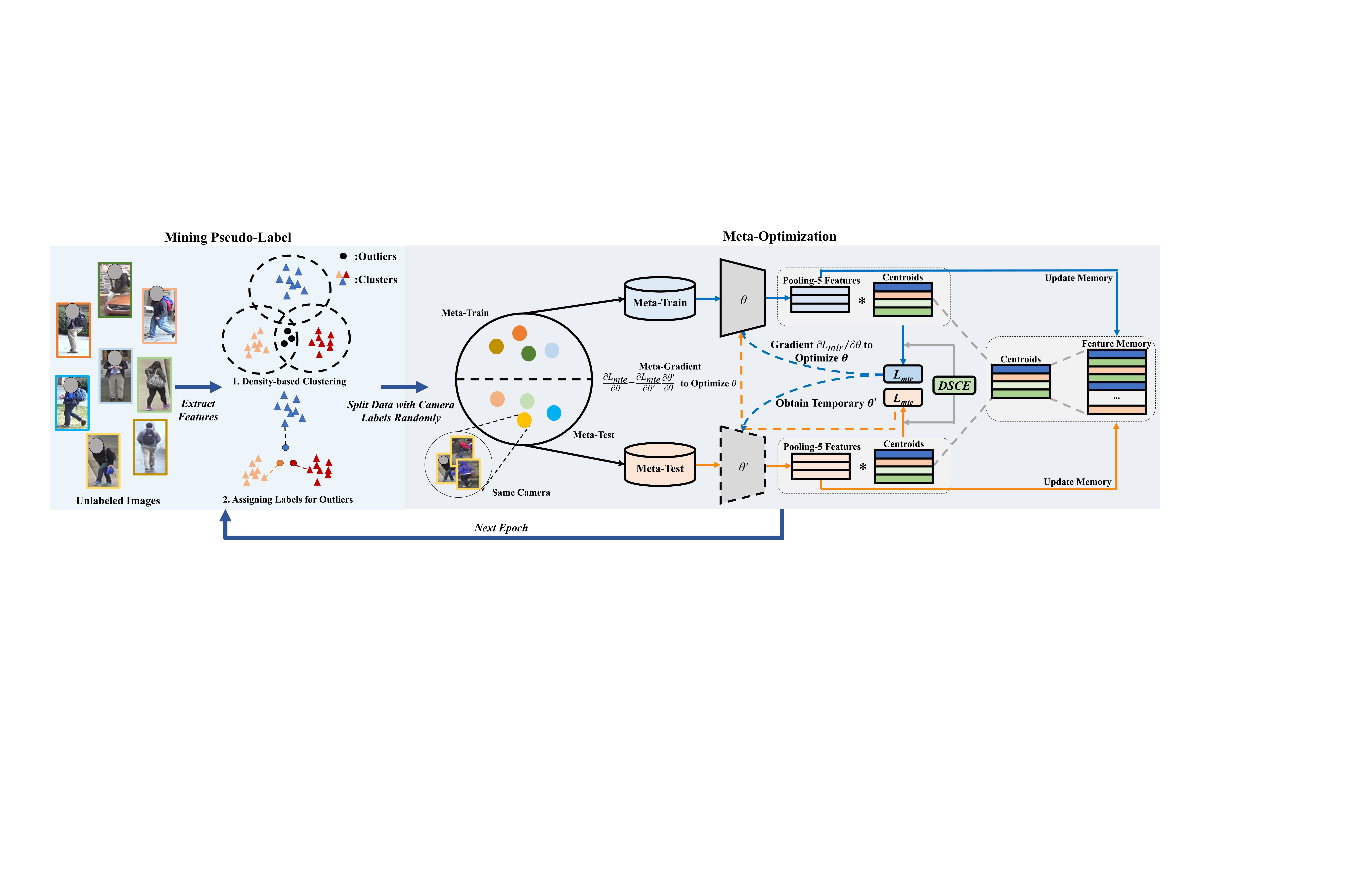}
    \caption{The framework for unsupervised re-ID, which includes two training stages, \ie, ``Mining Pseudo-Label" stage and ``Meta-Optimization". The first stage assigns samples with pseudo-label based on DBSCAN~\cite{ester1996density}. The second stage splits the training data into meta-train and meta-test sets based on camera labels and optimizes the model with the proposed meta-learning strategy. This camera-aware meta-learning (MetaCam) encourages the model to learn camera-invariant features. To reduce the negative impact of noisy labels, we also propose a dynamic and symmetric cross-entropy loss (DSCE) that is used for both meta-train and meta-test data. In our framework, the memory module saves the features of all samples, which enables us to dynamically build class centroids and thus to be adaptable to the change of clusters.
    }
    \vspace{-0.2in}
    \label{fig:framework}
\end{figure*}

\section{Related Work}
\subsection{Unsupervised Person Re-ID}
Unsupervised person re-ID can be categorized into \textit{Fully Unsupervised Re-ID} (FU)~\cite{lin2019bottom, lin2020unsupervised, zeng2020hierarchical} and \textit{Unsupervised Domain Adaptation} (UDA)~\cite{zhong2019invariance, yu2020weakly, li2019cross, qi2019novel}. The former tries to train a re-ID model with only unlabeled data while the latter attempts to leverage labeled source data and unlabeled target data to train a model for the target domain.
Although training under different data conditions, most methods for UDA and FU adopt similar learning strategies, which can be summarized into methods based on pseudo-label discovery~\cite{fu2019self, ge2020mutual, lin2019bottom, zeng2020hierarchical} and methods based on alignment~\cite{deng2018image, PTGAN, qi2019novel, zou2020joint}. The methods based on pseudo-label discovery rely on the iteration of pseudo-label mining and model fine-tuning, such as BUC~\cite{lin2019bottom}, SSG~\cite{fu2019self}, HCT~\cite{zeng2020hierarchical} and SpCL~\cite{ge2020selfpaced}. Despite their success, these methods might suffer from noisy samples obtained in the pseudo-label mining process. The alignment-based methods try to align distribution shift (\eg, camera shift or domain shift) in image level or feature level. In FU, we only have one dataset, therefore we mainly focus on reducing the camera shift in dataset. Zhong~\etal~\cite{Zhong_2018_ECCV} propose to align camera shift of the target samples by camera-to-camera style transfer. Yu~\etal~\cite{yu2020weakly} attempt to align the camera shift with 2-Wasserstein distance. The former is sensitive to distorted images in the generation process while the latter should be implemented under the precondition that features of pedestrians satisfy Gaussian distribution. Different from these two works, this paper aligns the camera shift by simulating the cross-camera process during training, instead of generating virtual images or learning based on prior assumptions.

\subsection{Learning with Noisy Labels}
Recent studies on learning with noisy labels (LNL) can mainly be categorized into sample re-weighting methods~\cite{shu2019meta, cheng2020learning, han2018co, yang2020asymmetric}, label correction methods~\cite{lee2018cleannet, vahdat2017toward, xiao2015learning} and robust loss designing methods~\cite{ghosh2017robust, wang2019symmetric, zhang2018generalized}. Re-weighting methods assign clean samples with higher weights during the training process. Shu~\etal~\cite{shu2019meta} attempt to train an additional network with meta-learning for sample re-weighting. However, it requires additional clean data to be meta-test set during optimization, which may not be available in real-world applications. Co-teaching~\cite{han2018co} proposes to remove noisy samples with a pair of networks. It can also be regarded as re-weighting methods, since the removed samples can be regarded as being assigned with zero weights. Label correction methods~\cite{lee2018cleannet, vahdat2017toward, xiao2015learning} try to identify noisy samples and correct their labels during training. These methods also require additional clean data to assist the model to correct labels. For robust loss functions, Ghosh~\etal~\cite{ghosh2017robust} propose a theory to check the robustness of loss functions and find that mean absolute error loss (MAE) is robust to noisy labels. However, the gradient saturation of MAE prevents it from obtaining satisfactory performance. To address this problem, Wang~\etal~\cite{wang2019symmetric} propose a symmetric cross-entropy loss for robust learning. Our work leverages the finding of \cite{ghosh2017robust,wang2019symmetric} and proposes a dynamic and symmetric cross-entropy loss to resist noisy labels produced by clustering.

\subsection{Meta Learning}
Meta-learning aims to learn new tasks with limited training samples and can be classified into optimizing-based~\cite{finn2017model, nichol2018reptile, rajeswaran2019meta}, model-based~\cite{santoro2016meta, munkhdalai2017meta} and metric-based~\cite{sun2018beyond, snell2017prototypical, sung2018learning} methods. Our work is closely related to optimizing-based meta-learning methods, which try to obtain a good initialized weight for fast adaptation for new tasks. Finn~\etal~\cite{finn2017model} propose model-agnostic meta-learning (MAML) to acquire the ideal weight by stimulating the learning process of new tasks with meta-test set. Subsequently, Reptile~\cite{nichol2018reptile} speeds up the learning process of MAML with a first-order approximation. Recently, MAML has been widely used in computer vision tasks such as noisy label learning~\cite{shu2019meta}, domain generalization~\cite{li2017learning} and face recognition~\cite{guo2020learning}. For the line of unsupervised learning, Hsu~\etal~\cite{hsu2018unsupervised} attempt to solve the few-shot problem with unlabeled meta-train data and labeled meta-test data.  Different from the above works, we implement meta-learning in the context of fully-unsupervised re-ID, where no labeled data are provided. In addition, this paper adopts the meta-learning to overcome the domain shift in re-ID, considering a completely different problem compared to existing meta-learning methods.

\section{Methodology}
\heading{Preliminary}. 
In the unsupervised person re-ID, we are provided with an unlabeled dataset $\mathcal{U} = \{x_1, x_2,..., x_N\}$ with $N$ images captured by $N_{cam}$ cameras. Generally, the distributions of images from different cameras will vary greatly.
The goal is to learn a model $\mathcal{F}$ parameterized by $\theta$ with $\mathcal{U}$, which can extract discriminative person feature $\mathbf{f} \in \mathbb{R}^{d \times 1}$ for retrieval.

\subsection{The Overall Framework}
Our overall framework is presented in Fig.~\ref{fig:framework}, which can be summarized into iterations of two stages: \textit{Mining Pseudo-Label} and \textit{Meta-Optimization}. In stage one, we first extract features for all training samples and then use DBSCAN~\cite{ester1996density} to assign pseudo-label for them. In stage two, we aim to use the generated pseudo-label to train the model. Specifically, we maintain a memory $\mathcal{W}$, which saves the features of all samples and is dynamically updated during training. The memory enables us to effectively train the model with changing pseudo-label obtained from the first stage. During training, we split the dataset $\mathcal{U}$ into meta-train set $\mathcal{M}_{tr}$ and meta-test set $\mathcal{M}_{te}$ based on camera labels. The model is trained with our proposed camera-aware meta-learning (MetaCam) strategy, which encourages the model to learn camera-invariant features. In addition, we propose a dynamic and symmetric cross-entropy loss (DSCE) for resisting noisy labels. These two stages are repeatedly iterated till the model converges.

\subsection{Mining Pseudo-Label}
To enable training on the unlabeled dataset $\mathcal{U}$, this paper adopts a popular way for generating pseudo-label, \ie, clustering-based strategy~\cite{song2020unsupervised, wang2020unsupervised}. Specifically, we first extract pooling-5 features for $\mathcal{U}$ by a re-ID model, which is initially pre-trained with ImageNet and will be updated in the training process. Given the extracted features, we compute their pair-wise Euclidean distances and calculate their Jaccard distances with $\emph{k}$-reciprocal nearest neighbours~\cite{zhong2017re}. The obtained Jaccard distances are used to generate pseudo-label for $\mathcal{U}$ with DBSCAN~\cite{ester1996density}. Since DBSCAN is a density-based clustering algorithm, it only assigns pseudo-label to high-confident samples (inliers) and remains low-confident samples as outliers. To fully utilize training samples in $\mathcal{U}$, we assign outliers with pseudo-label based on their corresponding nearest neighbours. Based on the above process, we produce pseudo-label for the unlabeled samples, which can be used for model optimization. However, due to the poor model initialization and camera variations, intra-identity samples might be assigned with different pseudo-label and inter-identity samples might be assigned with the same pseudo-label. Training with such noisy labels undoubtedly will hamper the model optimization and thus reduce the model performance. To address this problem, we propose DSCE loss and MetaCam for robust learning and overcoming the camera variations.

\subsection{Training with DSCE Loss}
Clustering-based unsupervised re-ID largely depends on the iteration between clustering and model optimizing stage~\cite{song2020unsupervised, zeng2020hierarchical}. There are two challenges in this process. (1) The number of centroids may change after each iteration, hindering the utilization of traditional cross-entropy loss that requires a fixed number of identities. (2) Clustering algorithm may bring a large amount of noisy samples in both inliers and outliers, which hurts model optimization. 

For the first challenge, inspired by ~\cite{li2020prototypical}, we propose a dynamic cross-entropy loss, which can be effectively utilized against the changing of centers. Specifically, we maintain a memory $\mathcal{W}$ that saves features for all training samples. During training, we construct online centroids from the memory by directly averaging over memory features that assigned with the same pseudo-label. The dynamic cross-entropy loss is formulated as,

\begin{equation}
    \label{eq:dce}
    L_{dce}(\mathbf{f}_{i}; \theta) = - \mathbf{\hat{\mathbf{y}}_{i}}^{\mathrm{T}} \log{\Big[\rm{Softmax}(\mathbf{C}^{\mathrm{T}} \mathbf{f}_{i} / \tau)\Big]},
\end{equation}
where $\mathbf{C} \in \mathbb{R}^{N_{c} \times d}$ represents the feature centers of each pseudo identity, $N_{c}$ is the number of clusters, $d$ is the feature dimension, and $\rm{Softmax}(\cdot)$ is the element-wise softmax function. $\mathbf{f}_{i}$ is the feature of $i$-th training sample extracted by the current model. $\mathbf{\hat{\mathbf{y}}_{i}} \in \mathbb{R}^{N_{c} \times 1}$ is the one-hot vector indicating the pseudo identity of $i$-th sample. 

For the second challenge, we aim to design a robust loss function for resisting noisy labels. Ghosh~\etal~\cite{ghosh2017robust} propose a theory to check whether a loss function $L$ is robust to noisy samples, which can be formulated as:
\begin{equation}
    \label{eq:theory}
    \sum_{k=1}^{N_c} L(\mathbf{f}, k) = Z,
\end{equation}
where $N_{c}$ is the number of categories and $Z$ is a constant. The above formula indicates that for any sample $\mathbf{f}$ and loss function $L$, the sum of losses about classifying $\mathbf{f}$ to all categories (\ie, 1 to $N_{c}$) should be a constant if $L$ is noise-tolerant. By utilizing this theory and drawing the inspiration from ~\cite{wang2019symmetric}, we introduce a dynamic and symmetric cross-entropy loss (DSCE) as:

\begin{equation}
    \label{eq:sym}
    L_{dsce}(\mathbf{f}_{i}; \theta) = - \Big[\rm{Softmax}(\mathbf{C}^{\mathrm{T}} \mathbf{f}_{i} / \tau) \Big]^{\mathrm{T}} \log{ \Big[ \rm{Softmax}(\hat{\mathbf{y}}_{i}) \Big] }.
\end{equation}

Different from~\cite{wang2019symmetric}, we adopt the softmax normalization to avoid the computational problem brought by $\log 0$ in one-hot vector $\hat{\mathbf{y}}_{i}$. The proposed $L_{dsce}$ utilizes a memory bank to adapt to the changing clusters in unsupervised re-ID and it also satisfies Eq.~\ref{eq:theory} (see~\appref{app:dsce} for details). Considering the good convergence of $L_{dce}$, the combined loss for optimization is:
\begin{equation}
    \label{eq:robust}
    L_{c}(\mathbf{f}_{i}; \theta) = L_{dce}+L_{dsce}.
\end{equation}

After each backpropagation step, we update the feature memory $\mathcal{W}$ with the following rule:
\begin{equation}
    \label{eq:memory}
    \mathcal{W}[i] = \alpha \mathcal{W}[i] + (1-\alpha) \mathbf{f}_{i},
\end{equation}
where $\alpha \in [0, 1]$ is the updating rate.

\subsection{Camera-Aware Meta-Learning}
The previous training scheme offers a basic solution to unsupervised re-ID, but it ignores the impact of camera shift, which is crucial for optimizing a robust re-ID model. The appearance of pedestrians under different cameras may suffer from the changes of viewpoint, illumination, and other environmental factors, leading to a large gap between intra-class features. Without considering this phenomenon, the trained model might be sensitive to the camera variations, which may decrease the clustering results and thus hampers the model optimization. To address this problem, we propose a camera-aware meta-learning strategy (MetaCam) to help the model learn a camera-invariant representation, which includes the following four steps: \textit{Meta-Sets Preparation}, \textit{Meta-Train}, \textit{Meta-Test}, and \textit{Meta-Update}.

\textbf{Meta-Sets Preparation}. In the proposed MetaCam, we aim to align the camera shift by simulating the cross-camera constraint during training. Given the training samples collected from $N_{cam}$ cameras, we split the training set into the meta-train set and meta-test set based on the camera labels. Specifically, we randomly select samples of $N_{mtr}$ cameras as the meta-train set $\mathcal{M}_{tr}$ and regard the samples of the rest $N_{cam} - N_{mtr}$ cameras as the meta-test set $\mathcal{M}_{te}$. Next, we will introduce how to utilize the generated meta-sets to learn a camera-invariant model.

\textbf{Meta-Train}. We calculate the meta-train loss on the mini-batch examples $m_{tr}$ sampled from $\mathcal{M}_{tr}$ with the proposed loss $L_{c}$ in Eq.~\ref{eq:memory}, formulated as:

%
\begin{equation}
    \label{eq:mtr}
    L_{mtr}(\mathcal{F}(m_{tr}); \theta) = \frac{1}{N_{b}} \sum_{i=1}^{N_b} L_{c} (\mathbf{f}_{i}; \theta),
\end{equation}
where $N_b$ is the batch size. By updating model parameters $\theta$ with SGD optimizer, we obtain a temporary model parameterized by $\theta^{'}$ for further optimization in the \textit{Meta-Test} step. The temporary model is obtained by:
\begin{equation}
    \label{eq:sgd}
    \theta^{'} = \theta - \gamma \frac{\partial L_{mtr}}{\partial \theta},
\end{equation}
where $\gamma$ is the learning rate. 

\begin{algorithm}[!t]
    \caption{The training procedure of proposed method.}
    \label{alg:metacam}
    \textbf{Inputs:} Unlabeled data $\mathcal{U}$ captured by $N_{cam}$ cameras, batch size $N_b$, re-ID model $\mathcal{F}$ parameterized by $\theta$, feature memory $\mathcal{W}$, number of meta-train cameras $N_{mtr}$, training epoch $epoch$, learning rate $\gamma$, updating rate $\alpha$. \\
    \textbf{Outputs:} Optimized model $\mathcal{F}$ parameterized with $\theta^{*}$.\\
    \vspace{-.15in}
    \begin{algorithmic}[1]
        \STATE Initialize $\mathcal{W}$ with $\mathbf{0}$;
        \FOR{$i$ in $epoch$}
            \STATE  \textcolor{gray}{// Stage 1: Mining Pseudo-Label.}
            \STATE Generate pseudo-label for $\mathcal{U}$ with DBSCAN;
            \STATE  \textcolor{gray}{// Stage 2: Meta-Optimization.}
            \STATE \textcolor{gray}{// Step 1: Meta-sets Preparation.}
            \STATE Select samples from $N_{mtr}$ random cameras as $\mathcal{M}_{tr}$ and regard samples of the rest cameras as $\mathcal{M}_{te}$.
            \REPEAT
                \STATE Sample mini-batch with $N_b$ meta-train samples $m_{tr}$ and $N_b$ meta-test samples $m_{te}$ .
                \STATE \textcolor{gray}{//Step 2: Meta-Train.}
                \STATE Compute meta-train loss on $m_{tr}$ with Eq.~\ref{eq:mtr};
                \STATE Obtain temporary $\theta^{'}$ with Eq.~\ref{eq:sgd};
                \STATE  \textcolor{gray}{//Step 3: Meta-Test.}
                \STATE Compute meta-test loss on $m_{te}$ with Eq.~\ref{eq:mte};
                \STATE \textcolor{gray}{// Step 4: Meta-Update.}
                \STATE Compute combined loss with Eq.~\ref{eq:meta};
                \STATE Update $\theta$ with gradient computed by Eq.~\ref{eq:grad}:
                \STATE Update $\mathcal{W}$ with $m_{tr}$ and $m_{te}$ based on Eq.~\ref{eq:memory}; 
            \UNTIL{$\mathcal{M}_{tr}$ and $\mathcal{M}_{te}$ are enumerated;}
        \ENDFOR
        \STATE $\theta^{*} \leftarrow \theta$.
        \STATE \textbf{Return} $\mathcal{F}$ parameterized with $\theta^{*}$.
    \end{algorithmic}
\end{algorithm}

\textbf{Meta-Test}. In the meta-test step, we aim at validating the accuracy of the temporary model $\theta^{'}$ on meta-test samples. To achieve this goal, we sample a mini-batch with $N_b$ images from $\mathcal{M}_{te}$ and compute the meta-test loss, formulated as:
\begin{equation}
    \label{eq:mte}
    L_{mte}(\mathcal{F}(m_{te}); \theta^{'}) = \frac{1}{N_{b}} \sum_{i=1}^{N_b} L_{c}(\mathbf{f}_{i}; \theta^{'}).
\end{equation}

\textbf{Meta-Update}. In this step, we update the model  with the combination of  meta-train loss and meta-test loss, which can be written as:
\begin{equation}
    \label{eq:meta}
    L_{meta}(\mathcal{F}(m_{tr}), \mathcal{F}(m_{te}); \theta) = L_{mtr} + L_{mte}.
\end{equation}

It should be noted that although the meta-test loss is computed based on temporary model $\theta^{'}$, the derivative \textit{w.r.t.} $\theta$ can also be obtained with the chain rule. Specifically, the derivative of $L_{meta}$ \textit{w.r.t.} the $\theta$ can be formulated as:
\begin{equation}
    \label{eq:grad}
    \frac{\partial L_{meta}}{\partial \theta} = \frac{\partial L_{mtr}}{\partial \theta} + \frac{\partial L_{mte}}{\partial \theta^{'}} \frac{\partial \theta^{'}}{\partial \theta}.
\end{equation}

To sum up, the overall training process of our method is listed in Alg.~\ref{alg:metacam}.

\begin{table*}[!t]
    \centering
    \caption{Comparison with state-of-the-arts (fully unsupervised). Our method out performs current unsupervised re-ID algorithms. ``*": Reproduced by~\cite{ding2020adaptive}, ``\dag": Reproduced based on the authors' code.}
    \label{tab:cmpSota}
    \resizebox{\linewidth}{!}{
        \begin{tabular}{l|c|ccc|ccc|ccc} 
            \hline
            \multirow{2}{*}{Methods} & 
            \multirow{2}{*}{Venue} &
            \multicolumn{3}{c}{DukeMTMC-reID} \vline &
            \multicolumn{3}{c}{Market-1501} \vline & 
            \multicolumn{3}{c}{MSMT-17}\\
            \cline{3-11}
             & & mAP & rank-1 & rank-5 & mAP & rank-1 & rank-5 & mAP & rank-1 & rank-5  \\
            \hline
             OIM~\cite{xiao2017joint} & CVPR'17 & 11.3 & 24.5 & 38.8 & 14.0 & 38.0 & 58.0 & - & - & -  \\
             BUC~\cite{lin2019bottom} & AAAI'19 & 27.5 & 47.4 & 62.6 & 38.3 & 66.2 & 79.6  & 3.4* & 11.5* & 18.6*  \\
             SSL~\cite{lin2020unsupervised} & CVPR'20 & 28.6 & 52.5 & 63.5 & 37.8 & 71.1 & 83.8 & - & - & -  \\
             MMCL~\cite{wang2020unsupervised} & CVPR'20 & 40.2 & 65.2 & 75.9 & 45.5 & 80.3 & 89.4 & - & - & - \\
             HCT~\cite{zeng2020hierarchical} & CVPR'20 & 50.7 & 69.6 & 83.4 & 56.4 & 80.0 & 91.6 & - & - & -  \\
             \hline
             ECN$^\dag$~\cite{zhong2019invariance} & CVPR'19 & 24.5 & 49.0 & 61.7 & 30.3 & 63.5 & 79.0 & 3.1 & 10.2 & 15.5 \\
             AE~\cite{ding2020adaptive} & TOMM'20 & 39.0 & 63.2 & 75.4 & 54.0 & 77.5 & 89.8 & 8.5 & 26.6 & 37.0  \\
             WFDR$^\dag$~\cite{yu2020weakly} & CVPR'20 & 42.4 & 62.0 & 75.1 & 50.1 & 72.1 & 80.5 & 8.6 & 22.3 & 32.5  \\
             Ours & This work & \bf 53.8 & \bf 73.8 & \bf 84.2 & \bf 61.7 & \bf 83.9 & \bf 92.3 & \bf 15.5 & \bf 35.2 & \bf 48.3  \\
            \hline
        \end{tabular}
    }
\end{table*}

\begin{table}[!t]
    \centering
    \renewcommand\arraystretch{1.2}
    \caption{Ablation study on the proposed method. ``Outliers": Including outliers into training data. ``DSCE": training with DSCE loss. ``MetaCam": training with MetaCam.}
    \label{tab:ablation}
    \resizebox{\linewidth}{!}{
    \centering
        \begin{tabular}{c|c|c|c|cc|cc} 
            \hline
            \multirow{2}{*}{No.} & 
            \multicolumn{3}{c}{Attributes} \vline & 
            \multicolumn{2}{c}{DukeMTMC-reID} \vline & 
            \multicolumn{2}{c}{Market-1501}  \\
            \cline{2-8}
             & Outliers & DSCE & MetaCam & mAP & rank-1 & mAP  & rank-1  \\
            \hline
             1 & $\times$ & $\times$ & $\times$ & 6.8 & 16.6 & 6.6 & 17.5 \\
             2 & $\checkmark$ & $\times$ & $\times$ & 39.2 & 59.7 & 51.2 & 73.2 \\
             3 & $\checkmark$ & $\checkmark$ & $\times$ & 43.4 & 62.8 & 53.9 & 74.8 \\
             4 & $\checkmark$ & $\times$ & $\checkmark$ & 51.1 & 71.2 & 59.4 & 82.1 \\
             5 & $\checkmark$ & $\checkmark$ & $\checkmark$ & \bf 53.8 & \bf 73.8 & \bf 61.7 & \bf 83.9 \\
            \hline
        \end{tabular}
    }
\end{table}

\textbf{Remark}. From Eq.~\ref{eq:grad}, we can observe that the proposed MetaCam encourages the model to be optimized to the direction that can perform well not only on samples from meta-train cameras but also on samples from meta-test cameras. The meta-test loss can be considered as a regularization term, which can lead the model to produce discriminative representations with high-order gradients.

\section{Experiments}
\label{sec:exp}
\heading{Datasets and Evaluation Protocol}. We evaluate our method on the three large-scale re-ID benchmarks,~\ie, Market-1501 (Market)~\cite{zheng2015scalable}, DukeMTMC-reID (Duke)~\cite{ristani2016MTMC,zheng2017unlabeled} and MSMT-17 (MSMT)~\cite{PTGAN}. Market includes $32,668$ images from $1,501$ persons under six cameras. Duke is composed of $36,411$ labeled images of $1,404$ identities from eight cameras. MSMT has $126,441$ samples from $4,101$ pedestrians captured by fifteen cameras. For each dataset, nearly half of the identities are used for training and the remaining identities are used for testing. We adopt mAP and rank-1/5 accuracy for evaluation.

\heading{Implementation Details}. We adopt the ResNet-50~\cite{he2016deep} as the backbone. The ``exemplar-invariance" constraint in ECN~\cite{zhong2019invariance} is used to initialize our model and memory for 5 epochs. In our method, the number of cameras in the meta-train set $N_{mtr}$ are set to 3, 4 and 7 for Market, Duke and MSMT, respectively. During training, we set the learning rate $\gamma=3.5 \times 10^{-4}$, batch size $N_b=64$, temperature factor $\tau=0.05$, updating rate $\alpha=0.2$. Images are resized to $256 \times 128$. We use random crop, random flip and random erasing~\cite{zhong2020random} for data augmentation. The model is updated by the Adam optimizer. We train the model with 40 epochs in total, \ie, $max\_epoch=40$. During testing, we extract $2048$-dim pooling-5 features for retrieval.

\subsection{Comparison with State-of-the-Art}
We evaluate our method on Market, Duke and MSMT and compare it with state-of-the-art methods: including OIM~\cite{xiao2017joint}, BUC~\cite{lin2019bottom}, SSL~\cite{lin2020unsupervised}, MMCL~\cite{wang2020unsupervised}, HCT~\cite{zeng2020hierarchical}, ECN~\cite{zhong2019invariance}, AE~\cite{ding2020adaptive} and WFDR~\cite{yu2020weakly}. To fairly compare our MetaCam with WFDR~\cite{yu2020weakly} that aligns the camera feature shift with 2-Wasserstein distance, we implement WFDR in our framework by replacing MetaCam with WFDR. We also reproduce ECN~\cite{zhong2019invariance}, which is an unsupervised domain adaptation method and considers the camera shift, based on the provided source code. From Tab.~\ref{tab:cmpSota}, we make the following two conclusions. 
(1) Our method achieves the best results on three large-scale datasets. Specifically, we achieve \textbf{mAP=53.8\%} and \textbf{rank-1 accuracy=73.8\%} for Duke, \textbf{mAP=61.7\%} and \textbf{rank-1 accuracy=83.9\%} for Market, and \textbf{mAP=15.5\%} and \textbf{rank-1 accuracy=35.2\%} for MSMT. Compared to the currently best published method HCT~\cite{zeng2020hierarchical}, our method surpasses it by 3.1\% on Duke and 5.3\% on Market in mAP. This demonstrates that our method produces the new state of the art result for unsupervised person re-ID.
(2) Compared to methods (ECN~\cite{zhong2019invariance}, AE~\cite{ding2020adaptive}, and WFDR~\cite{yu2020weakly}) that consider the camera variations during model training, our method produces significantly higher results. Specifically, when using the same framework, our method (w/ MetaCam) clearly outperforms WFDR~\cite{yu2020weakly} in all datasets. This demonstrates the effectiveness of the proposed MetaCam in addressing the camera shift for unsupervised re-ID.

\begin{figure*}[!t]
    \centering
    \includegraphics[width=0.9\linewidth]{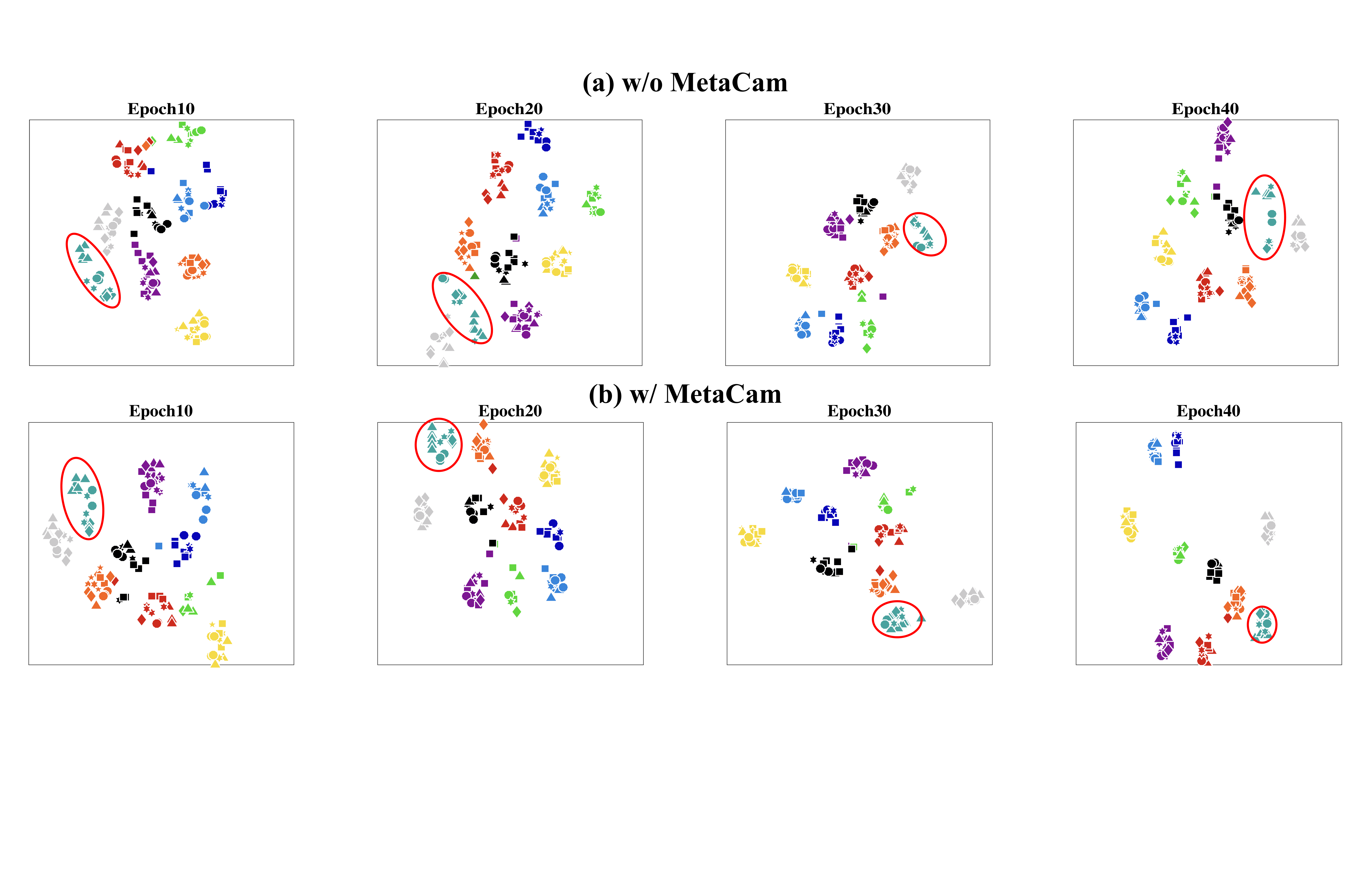}
    \caption{\emph{t}-SNE plot of 10 persons under different settings (model trained w/o MetaCam and model trained w/ MetaCam). We use different colors to denote identities and different shapes to indicate camera IDs. The algorithm with MetaCam generates intra-class features that are close to each other, indicating that our MetaCam can guide the model to learn camera-invariant features.}
    \label{fig:tsne}
    \vspace{-.1in}
\end{figure*} 

\begin{figure}[!t]
    \centering
    \includegraphics[width=0.95\linewidth]{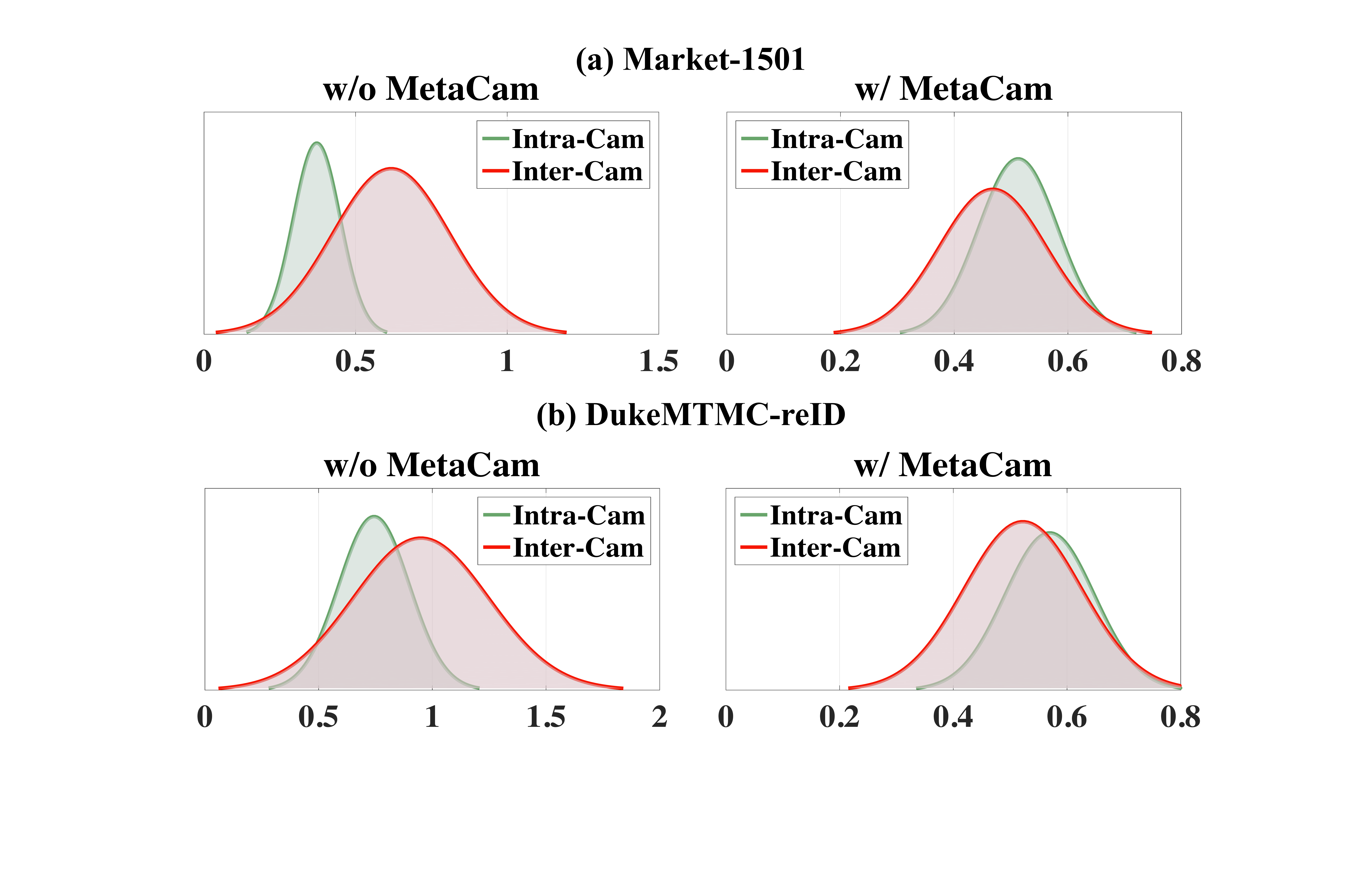}
    \caption{Distance distributions of positive pairs for intra-camera (\textGreen{green} curve) and inter-camera (\textRed{red} curve). We compare the results between the model trained with or without MetaCam on Market and Duke. }
    \label{fig:hist}
    \vspace{-.2in}
\end{figure} 

\subsection{Ablation Study}

We conduct experiments to investigate three important components of our methods, \ie 1) using outliers that assign labels for low-confident samples generated by clustering, 2) DSCE loss that is designed to prevent the model from over-fitting on noisy samples, and 3) MetaCam that is proposed to overcome camera variations. Ablative experiments on these three components are reported in Tab.~\ref{tab:ablation}.

\heading{The effectiveness of using outliers}. Without using these three components, the model achieves poor results on both datasets. The main reason is that DBSCAN will regard most of the samples as outliers with features extracted by the poor initial model, leading to the limited training samples during optimization and thereby produces undesired results.
When using outliers during training, the results are significantly improved. This demonstrates the importance of assigning pseudo-label for outliers and using them during model training. In the following experiments, we use outliers in training by default.

\heading{The effectiveness of DSCE loss function}. When adding DSCE loss into the model, we can achieve consistent improvement, no matter whether to use MetaCam or not. This verifies the advantage of DSCE loss when training the model with noisy labels generated by the clustering step.

\heading{The effectiveness of MetaCam}. From the comparison of \textit{No. 2} vs \textit{No. 4} and \textit{No. 3} vs \textit{No. 5}, we obtain two observations. First, MetaCam can significantly improve the results, demonstrating the necessity of overcoming the camera variation in unsupervised re-ID and the effectiveness of the proposed MetaCam. Second, the proposed MetaCam and  DSCE loss are complementary to each other. When combining them, the model can gain more improvement in performance.

\subsection{Visualization for MetaCam}

To better understand the effect of our MetaCam in overcoming the camera variations, we conduct two visualization experiments: (1) \emph{t}-SNE~\cite{maaten2008visualizing} plot of feature embeddings with the evolution of training; (2) distance distribution of intra-camera and inter-camera samples for the same ID.

\textbf{\emph{t}-SNE plot of feature embeddings}. We randomly select 10 persons from Market and visualize their features with \emph{t}-SNE~\cite{maaten2008visualizing} in different training epochs. In Fig.~\ref{fig:tsne}, we show the results of the model trained with or without MetaCam, respectively. For fair comparison, both models use the DSCE loss during training. We use different colors to denote identities and different shapes to represent camera IDs. We can find two phenomenons. (1) Features of the same identity are progressively gathered with the model training for both settings. This demonstrates that our method is able to learn discriminative person representations. (2) The model trained with MetaCam can produce more compact feature clusters (\eg, dark green points highlighted by the red circle). The intra-class features under different cameras are well gathered with the help of MetaCam. This verifies the advantage of our MetaCam in learning camera-invariant features. In addition, with the camera-invariant features, we can generate more accurate pseudo-label in the clustering step, which can further facilitate the optimization.

\textbf{Distance distribution}. To more precisely investigate the influence of MetaCam, we conduct experiments to visualize the distance distribution of positive pairs for intra-camera and inter-camera.  Specifically, we randomly select 50,000 distances for both intra-camera pairs and inter-camera pairs and draw the histogram for each setting. Results compared between the model trained with or without MetaCam are illustrated in Fig.~\ref{fig:hist}.
We can make two observations. (1) The model trained without MetaCam leads to a large gap between intra-camera distribution and inter-camera distribution. Specifically, the distances between positive pairs of inter-camera are commonly larger than that of intra-camera, indicating that the model trained without MetaCam is sensitive to camera variations. (2) When training with MetaCam, the distribution gap between positive pairs for intra-camera and inter-camera is significantly reduced. Concretely, the distances between positive pairs of inter-camera are commonly similar to that of intra-camera. This suggests that our MetaCam is able to align the camera shift and can lead the model to learn camera-invariant features, which is an important factor in person re-ID.

\begin{figure}[!t]
    \centering
    \includegraphics[width=0.95\linewidth]{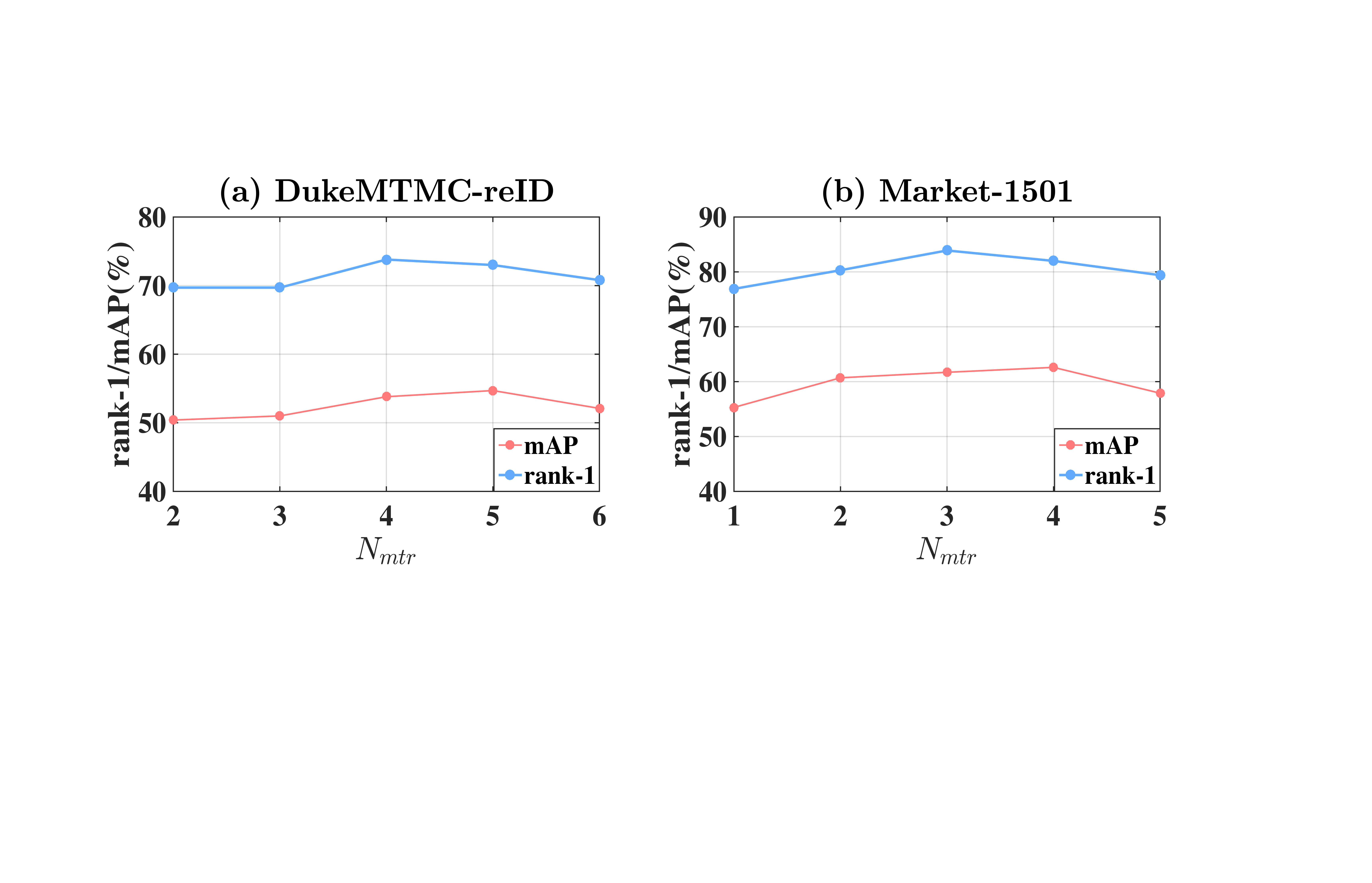}
    \vspace{-.05in}
    \caption{Sensitivity analysis of $N_{mtr}$.}
    \label{fig:sensitive}
\end{figure} 

\begin{figure}[!t]
    \centering
    \includegraphics[width=0.95\linewidth]{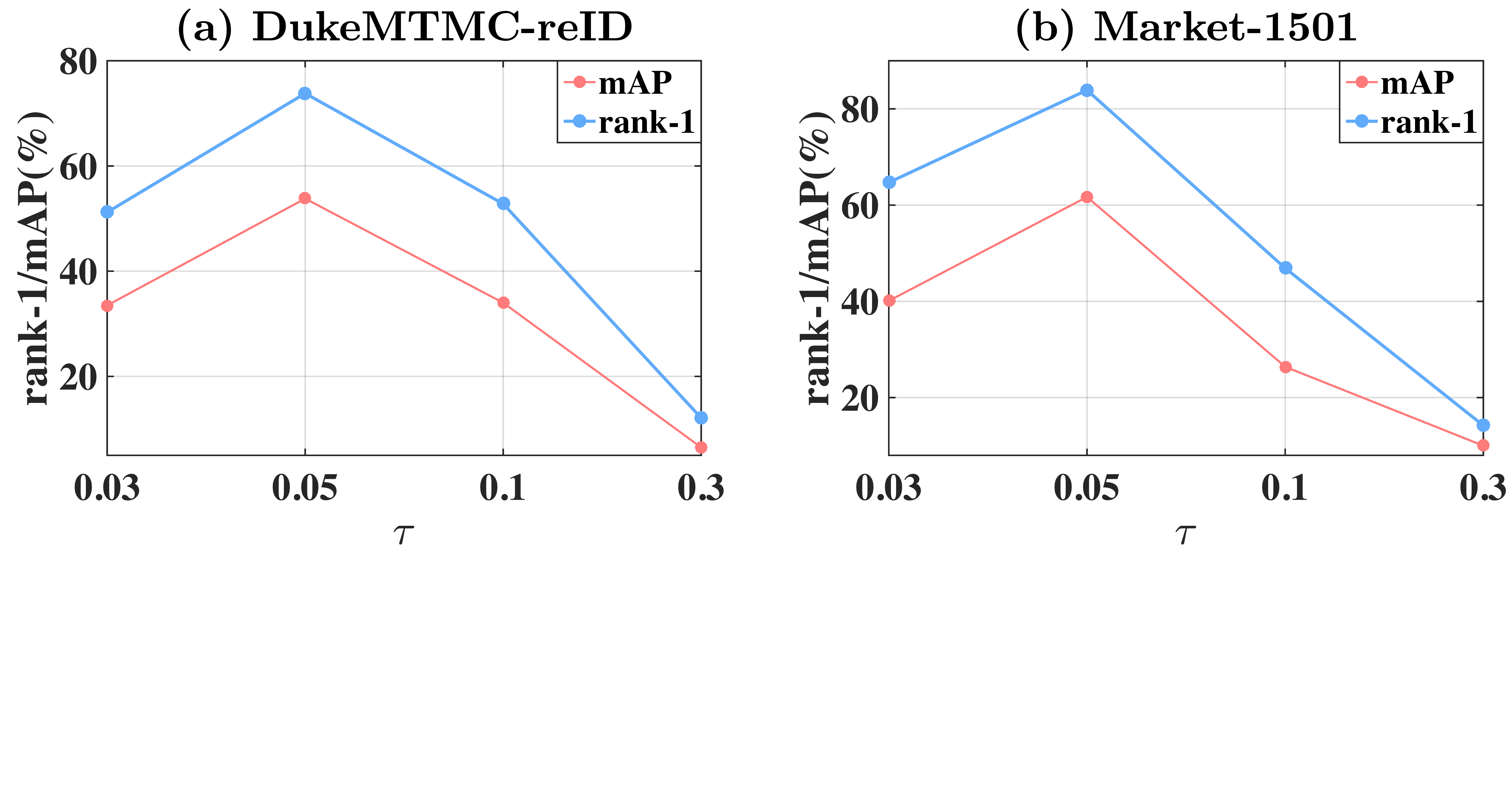}
    \vspace{-.05in}
    \caption{Sensitivity analysis of $\tau$.}
    \label{fig:temp}
    \vspace{-.2in}
\end{figure} 

\subsection{Sensitivity Analysis}
We further analyze the sensitivity to two hyper-parameters of our method, \ie, the number of cameras in meta-train $N_{mtr}$ for MetaCam and temperature factor $\tau$ for the memory-based loss. In our experiments, we change the value of one hyper-parameter and the others remain fixed.

\heading{Sensitivity to $N_{mtr}$}. In Fig.~\ref{fig:sensitive}, we vary $N_{mtr}$ from $1$ to $5$ for Market and $2$ to $6$ for Duke to investigate the effect of involving samples of different cameras into meta-train. 
We find that the best accuracy is achieved when $N_{mtr}$ is equal to half of the total number of cameras for each dataset. This indicates that it is better to keep a balance camera variations in the meta-train and meta-test.

\heading{Sensitivity to $\tau$}. In Fig.~\ref{fig:temp}, we investigate the effect of temperature factor $\tau$. A smaller $\tau$ leads to lower entropy and may help to achieve better results in re-ID. However, the $\tau$ with too small value, such as $0.03$, will cause the collapse of training. In our experiments, we set $\tau=0.05$, which achieves well performance across all datasets.

\subsection{Results for Domain Adaptation}
We also apply our method to the setting of unsupervised domain adaptation (UDA). In UDA, we are additionally given a labeled source domain, which can provide extra supervision for model training. Since our method is designed to learn the model with unlabeled data, we initialize the re-ID backbone with the model trained on MMT~\cite{ge2020mutual} and fine-tune the model with our method on the unlabeled data.
We evaluate our method on the settings of transferring between Duke and Market.
Comparisons with state-of-the-art methods are reported in Tab.~\ref{tab:uda}. All the compared methods use ResNet-50 as the backbone. We can observe that MMT~\cite{ge2020mutual} achieves the best results. After adding our method, the performance is further improved. Specifically, our method increases the mAP from 71.2\% to 76.5\% when testing on Market and from 63.1\% to 65.0\% when testing on Duke. This indicates that our method is also suitable for UDA and can be readily applied to further improve the performance of other UDA methods.

\begin{table}[!t]
    \centering
    \renewcommand\arraystretch{1.2}
    \caption{Results on domain adaptation. M: Market-1501, D: DukeMTMC-reID. MMT-500: MMT~\cite{ge2020mutual} with $\emph{k}=500$ for \emph{k}-means clustering. All methods use ResNet-50 as the backbone.}
    \label{tab:uda}
    \resizebox{\linewidth}{!}{
    \centering
        \begin{tabular}{c|c|cc|cc} 
            \hline
            \multirow{2}{*}{Methods} & 
            \multirow{2}{*}{Venue} & 
            \multicolumn{2}{c}{D $\rightarrow$ M} \vline & 
            \multicolumn{2}{c}{M $\rightarrow$ D}  \\
            \cline{3-6}
             &  & mAP & rank-1 & mAP  & rank-1  \\
            \hline
             SPGAN~\cite{deng2018image} & CVPR'18 & 22.8 & 51.5 & 22.3 & 44.1 \\
             HHL~\cite{Zhong_2018_ECCV} & ECCV'18 & 31.4 & 62.2 & 27.2 & 46.9 \\
             ECN~\cite{zhong2019invariance} & CVPR'19 & 43.0 & 75.1 & 40.4 & 63.3 \\
             SSG~\cite{fu2019self} & ICCV'19 & 58.3 & 80.0 & 53.4 & 73.0 \\
             UCDA-CCE~\cite{qi2019novel} & ICCV'19 & 34.5 & 64.3 & 36.7 & 55.4 \\
             MMCL~\cite{wang2020unsupervised} & CVPR'20 & 60.4 & 84.4 & 51.4 & 72.4 \\
             DG-Net++~\cite{zou2020joint} & ECCV'20 & 61.7 & 82.1 & 63.8 & 78.9 \\
             GDS~\cite{jin2020global} & ECCV'20 & 61.2 & 81.1 & 55.1 & 73.1 \\
             \hline
             MMT-500~\cite{ge2020mutual} & ICLR'20 & 71.2 & 87.7 & 63.1 & 76.8 \\
             MMT-500+Ours & This Work & \bf 76.5 & \bf 90.1 & \bf 65.0 & \bf 79.5 \\
            \hline
        \end{tabular}
    }

    \vspace{-0.2in}
\end{table}

\section{Conclusion}
In this paper, we propose a novel framework for unsupervised re-ID, which is designed based on a Dynamic and Symmetric Cross-Entropy loss (DSCE) and a camera-aware meta-learning algorithm (MetaCam). Our DSCE is able to handle the changing clusters and can resist noisy samples during model optimization. The proposed MetaCam can effectively reduce the camera shift by simulating the cross-camera searching process during training. Extensive experiments show the effectiveness of our method, which can achieve state-of-the-art results on three datasets for both fully-unsupervised re-ID and domain adaptive re-ID. 

\small{{\noindent\textbf{Acknowledgements}} This work is supported by the National Nature Science Foundation of China (No. 61876159, 61806172, 61662024, 62076116 and U1705286); the China Postdoctoral Science Foundation Grant (No. 2019M652257); the Fundamental Research Funds for the Central Universities (Xiamen University, No. 20720200030); the European Commission under European Horizon 2020 Programme (No. 951911 - AI4Media) and the  Italy-China collaboration project TALENT (No. 2018YFE0118400).}

\begin{appendices}

    \section{Detailed Explanation of DSCE }
                
        \label{app:dsce}
        We use $\mathbf{C}_{j}$ to denote the $j$-th centroid ($1 \leq j \leq N_c$). $p_{j}=\frac{\exp(\mathbf{C}_{j}^{\mathrm{T}} \mathbf{f} / \tau)}{\sum_{m=1}^{N_c} \exp(\mathbf{C}_{m}^{\mathrm{T}} \mathbf{f} / \tau)}$ is the probability of assigning $\mathbf{f}$ to the $j$-th class and $\sum_{j=1}^{N_c} p_j = 1$. $\hat{\mathbf{y}}$ is the one-hot vector of $\mathbf{f}$ obtained by clustering. $\widetilde{y}_j = \frac{\exp{(\hat{y}_j)}}{\sum_{m=1}^{N_c} \exp{(\hat{y}_m)}}$ is the $j$-th element of softmax-normalized $\hat{\mathbf{y}}$. Then, we have:
                
        \begin{equation}
            \label{eq:y}
                \widetilde{y}_j=\begin{cases}
                \frac{1}{N_c -1 + \textit{e}}, & \hat{y}_j=0 \\
                \frac{\textit{e}}{N_c -1+\textit{e}}, & \hat{y}_j=1
                \end{cases}.
        \end{equation}
        
        The DSCE loss in our paper (Eq.~\ref{eq:sym}) can be reformulated as:
        
        \begin{equation}
            \label{eq:dsce_re}
            L_{dsce} = -\sum_{j=1}^{N_c} p_j \log{\widetilde{y}_{j}}.
        \end{equation}
        
        \cite{ghosh2017robust} proves that a loss function $L$ is robust to noisy labels if it satisfies Eq.~\ref{eq:theory}, where $L(\mathbf{f}, k)$ indicates the loss when the class label is $k$. Next, we will prove that DSCE loss satisfies Eq.~\ref{eq:theory}.
                
        \begin{theorem}
            In a multi-class classification problem, the proposed DSCE loss (Eq.~\ref{eq:dsce_re}) satisfies the constraint in Eq.~\ref{eq:theory}. 
            \begin{proof}
            
                \rm When the class label is $k$-th class, \textit{i.e.,} $\hat{y}_k=1$, Eq.~\ref{eq:dsce_re} can be reformulated as:
                \begin{equation}
                    \label{eq:dscek}
                    L_{dsce}(\mathbf{f}, k) = - p_{k} \log(\widetilde{y}_{k}) - \sum_{j \neq k}^{N_c} p_{j} \log(\widetilde{y}_{j}).
                \end{equation}
    
                For convenience, we define $Q = \frac{1}{N_c -1 + \textit{e}}$. According to Eq.~\ref{eq:y}, Eq.~\ref{eq:dscek} can be simplified as:
    
                \begin{align}
                    L_{dsce}(\mathbf{f}, k) &= - p_{k} \log(\textit{e}Q) - (\log Q) \sum_{j \neq k}^{N_c} p_{j}  \notag \\
                                            &= - (1+\log Q)p_{k} -(1-p_{k})\log Q \notag \\
                                            &= -p_{k}-\log Q \notag.
                \end{align}
    
                Then, we have:
                \begin{align}
                    \sum_{k=1}^{N_c} L_{dsce}(\mathbf{f}, k) &= -\sum_{k=1}^{N_c} p_{k}-\sum_{k=1}^{N_c} \log Q \notag \\
                                                             &= -1-N_{c} \log Q \notag. 
                \end{align} 
                
                Therefore, the proposed DSCE loss satisfies Eq.~\ref{eq:theory} and is robust to noisy labels.
                $\hfill \blacksquare$
        \end{proof} 
    \end{theorem}

\end{appendices}

{\small
\bibliographystyle{ieee_fullname}
\bibliography{egbib}
}

\end{document}